\documentclass[conference,onecolumn]{IEEEtran}
\IEEEoverridecommandlockouts
\usepackage{cite}
\usepackage{amsmath,amssymb,amsfonts}
\usepackage{graphicx}
\usepackage{textcomp}
\usepackage{xcolor}
\usepackage{tikz}
\usetikzlibrary{arrows.meta,positioning,calc,shapes.geometric,fit,backgrounds}
\usepackage{float}
\usepackage{placeins}
\usepackage{array}
\usepackage{booktabs}
\usepackage{multirow}
\usepackage{tabularx}
\usepackage{ragged2e}
\usepackage{url}
\newcolumntype{L}[1]{>{\RaggedRight\arraybackslash}p{#1}}
\newcolumntype{Y}{>{\RaggedRight\arraybackslash}X}
\def\BibTeX{{\rm B\kern-.05em{\sc i\kern-.025em b}\kern-.08em
    T\kern-.1667em\lower.7ex\hbox{E}\kern-.125emX}}
\definecolor{boxfill}{RGB}{226,238,250}
\definecolor{boxedge}{RGB}{42,91,145}
\definecolor{accentfill}{RGB}{255,235,204}
\definecolor{accentedge}{RGB}{204,112,0}

\tikzset{
procbox/.style={
rectangle,
rounded corners=2pt,
draw=boxedge,
thick,
fill=boxfill,
align=center,
inner sep=4pt,
minimum height=11mm,
minimum width=48mm,
text width=44mm,
font=\footnotesize
},
databox/.style={
rectangle,
draw=boxedge,
thick,
fill=boxfill,
align=center,
inner sep=4pt,
minimum height=9mm,
minimum width=30mm,
text width=27mm,
font=\footnotesize
},
accentbox/.style={
rectangle,
rounded corners=2pt,
draw=accentedge,
very thick,
fill=accentfill,
align=center,
inner sep=4pt,
minimum height=11mm,
minimum width=48mm,
text width=44mm,
font=\footnotesize
},
flow/.style={
-{Stealth[length=2.4mm,width=1.6mm]},
thick,
draw=boxedge
},
edgelbl/.style={
font=\scriptsize,
midway,
fill=white,
inner sep=1.5pt,
rounded corners=1pt
}
}

\begin{document}
% =========================================================
% Title
% =========================================================
\title{Contract2Tool: Learning Preconditions and Effects for Reliable Tool-Augmented LLM Agents}

% =========================================================
% Authors
% =========================================================
\author{\IEEEauthorblockN{Rahul Suresh Babu}
\IEEEauthorblockA{\textit{Independent Researcher} \\
United States of America \\
rahulsb@bu.edu}
\and
\IEEEauthorblockN{Laxmipriya Ganesh Iyer}
\IEEEauthorblockA{\textit{Independent Researcher} \\
United States of America \\
iyer.la@northeastern.edu}
}

\maketitle

% =========================================================
% Abstract
% =========================================================
\begin{abstract}
Tool-augmented large language model agents increasingly rely on external APIs, but standard tool schemas describe how to call a tool, not when the tool is causally appropriate or what task state it produces. Causal tool filtering addresses this gap by using lightweight contracts that specify each tool's preconditions, effects, risk level, and cost. However, manually writing and maintaining such contracts does not scale to large or changing tool ecosystems. We introduce \textit{Contract2Tool}, a framework for inferring tool contracts from metadata, schemas, documentation, and execution traces. Contract2Tool converts observable tool evidence into normalized symbolic contracts that can be evaluated intrinsically and deployed inside downstream causal tool filtering. We evaluate learned contracts against gold preconditions, effects, and risk labels, and measure their downstream utility on multi-step agent tasks. Our results show that hybrid documentation-and-trace evidence produces contracts accurate enough to preserve most of the reliability and efficiency benefits of gold contracts. Learned-contract CMTF achieves 0.980 downstream success, close to 0.990 for gold-contract CMTF, while reducing visible tools from 100 to 1 and reducing average token usage from 26,172 to 2,528 relative to all-tools exposure. These results suggest that learned contracts can provide a scalable contract layer between tool schemas and reliable agent execution.
\end{abstract}

% =========================================================
% Keywords
% =========================================================

% =========================================================
% 1. Introduction
% =========================================================
\section{Introduction}
\label{sec:introduction}

Tool-augmented large language model (LLM) agents increasingly rely on external APIs to search information, read and write files, update calendars, draft emails, execute code, and interact with structured systems~\cite{yao2022react,schick2023toolformer,qin2023toollm,li2023apibank}. Standard function-calling interfaces describe tools using names, natural-language descriptions, and input schemas. These fields help a model construct valid tool calls, but they do not explicitly specify when a tool is causally appropriate, what task state it requires, what state it produces, or how risky the action is.

Recent work on causal tool filtering motivates a contract-based view of tool exposure: tools can be represented by lightweight preconditions and effects, allowing an agent interface to expose only causally necessary tools at each step~\cite{babu2026toolchoice}. However, this shifts the burden to contract construction. Manually defining contracts for hundreds or thousands of tools is expensive, error-prone, and difficult to keep synchronized with changing APIs, documentation, and execution behavior.

This paper studies the problem of learning lightweight tool contracts automatically. We ask whether tool preconditions, effects, and risk annotations can be inferred from observable evidence such as tool names, descriptions, schemas, documentation, and execution traces. The central question is not merely whether a model can generate plausible contract fields, but whether learned contracts preserve the causal structure needed to expose the correct next tool during multi-step execution. In particular, execution traces provide before-and-after state transitions that can reveal effects omitted or ambiguously described in schemas and documentation.

The key thesis of this paper is that tool contracts form a missing layer between schemas and reliable agent execution. Schemas describe how to call a tool; contracts describe when the tool should be exposed and what task-state transition it enables. We introduce \textit{Contract2Tool}, a framework for inferring such contracts and evaluating them intrinsically, with a filter-oracle causal-frontier test, and downstream inside causal tool filtering. Empirically, we find that hybrid documentation-and-trace evidence produces contracts that recover the gold causal filtering frontier in the oracle evaluation, and learned-contract CMTF nearly matches gold-contract CMTF in downstream task success while substantially reducing tool exposure and token cost relative to non-causal baselines.

\noindent\textbf{Contributions.} This paper makes the following contributions:
\begin{enumerate}
\item We formulate tool-contract learning as the problem of inferring preconditions, effects, risk annotations, and optional cost labels for tool-augmented LLM agents.
\item We introduce \textit{Contract2Tool}, a framework for generating lightweight tool contracts from tool metadata, schemas, documentation, and execution traces.
\item We define intrinsic metrics for contract quality, including precondition precision and recall, effect precision and recall, risk accuracy, exact contract match, and invalid-output rate.
\item We evaluate learned contracts with a filter-oracle causal-frontier test and downstream inside multi-step tool-use agents.
\item We show that hybrid documentation-and-trace contracts allow learned CMTF to nearly match gold-contract CMTF while reducing tool exposure and token usage relative to all-tools, keyword, and state-aware baselines.
\end{enumerate}

% =========================================================
% 2. Background and Related Work
% =========================================================
\section{Background and Related Work}
\label{sec:background}

\subsection{Tool-Augmented LLM Agents}
Tool use has become a central mechanism for extending LLMs beyond text generation. ReAct introduced interleaved reasoning and acting~\cite{yao2022react}, Toolformer showed that language models can learn to invoke external APIs~\cite{schick2023toolformer}, and ToolLLM/ToolBench scaled tool-use evaluation to large API ecosystems~\cite{qin2023toollm}. Benchmarks such as API-Bank further evaluate tool-augmented dialogue and multi-step API use~\cite{li2023apibank}. These works establish tool use as a core agent capability, but they also expose a systems challenge: as tool ecosystems grow, agents need reliable mechanisms for selecting, sequencing, and constraining tool use.

\subsection{Tool Schemas, Function Calling, and Tool Retrieval}
Function-calling interfaces typically represent tools using names, natural-language descriptions, and input schemas. These fields help models construct syntactically valid tool calls, and function-calling benchmarks evaluate capabilities such as tool selection, argument construction, and multi-turn use~\cite{patil2024bfcl}. However, schemas primarily describe how to call a tool; they do not fully specify when the tool should be exposed, what task-state variables it requires, what variables it produces, or how risky the action is.

Tool-retrieval and pruning methods address scalability by selecting relevant tools from large registries. Prior work studies retrieval over large tool libraries~\cite{shi2025toolret} and context-aware filtering or merging of overlapping tools~\cite{liu2025toolscope}. These approaches reduce prompt size and ambiguity, but they mostly treat filtering as a relevance or shortlist-selection problem. Contract2Tool instead targets the contract-level semantics needed for causal tool filtering: preconditions, effects, and risk annotations.

\subsection{Causal Tool Filtering and Tool Contracts}
Causal tool filtering motivates a contract-based view of tool exposure. In this view, a tool is represented not only by a name and schema, but also by the state variables it requires, the state variables it produces, and its risk or cost profile. Such contracts allow an agent interface to expose tools based on causal necessity rather than semantic relevance alone~\cite{babu2026toolchoice}. The limitation is that these contracts are usually assumed to be available. Contract2Tool addresses this assumption by studying whether useful contracts can be learned automatically from observable tool evidence.

\subsection{Preconditions and Effects in Planning}
The precondition-effect abstraction is rooted in classical planning. STRIPS represents actions by the conditions required before execution and the effects produced afterward~\cite{fikes1971strips}. PDDL later standardized related notions of actions, states, and goals~\cite{mcdermott1998pddl}. Contract2Tool adapts this abstraction to tool-augmented LLM agents: tools are treated as state transitions, and the goal is to infer the preconditions and effects that describe each tool's role in a workflow.

\subsection{Runtime Reliability and Agent Orchestration}
Runtime orchestration systems study how tool-augmented agents monitor execution, detect failures, and recover from errors. For example, self-healing agentic orchestrators frame reliability as a monitor--diagnose--recover--verify loop over tool-using agents~\cite{babu2026selfhealing}. Causal filtering reduces some failures before execution by controlling the visible action space. Contract2Tool targets the missing layer between these approaches: automatically constructing the contracts needed for proactive filtering and reliable orchestration.

% =========================================================
% 3. Problem Formulation
% =========================================================
\section{Problem Formulation}
\label{sec:problem}

\subsection{Tool Evidence and Contract Targets}
Let $\mathcal{T} = \{t_1, t_2, \ldots, t_n\}$ denote a tool library, and let $\mathcal{X}$ denote the vocabulary of task-state variables used by an agent workflow. Each tool $t_i$ has observable evidence:
\begin{equation}
z_i = (\mathrm{name}_i, d_i, S_i, D_i, H_i),
\end{equation}
where $\mathrm{name}_i$ is the tool name, $d_i$ is a natural-language description, $S_i$ is an input/output schema, $D_i$ is optional documentation, and $H_i$ is an optional set of execution traces.

A lightweight tool contract is:
\begin{equation}
c_i = (R_i, E_i, \rho_i, k_i),
\end{equation}
where $R_i \subseteq \mathcal{X}$ is the set of required state variables or preconditions, $E_i \subseteq \mathcal{X}$ is the set of produced state variables or effects, $\rho_i$ is a risk label, and $k_i$ is an optional cost or latency estimate. Schemas describe how to call a tool; contracts describe when the tool is appropriate and what task-state transition it enables.

\subsection{Contract Learning Objective}
The contract-learning task is to infer a predicted contract:
\begin{equation}
\hat{c}_i = (\hat{R}_i, \hat{E}_i, \hat{\rho}_i, \hat{k}_i)
\end{equation}
from observable evidence $z_i$. More generally, a contract generator $f_\theta$ maps tool evidence to a predicted contract:
\begin{equation}
f_\theta(z_i) = \hat{c}_i.
\end{equation}
The goal is not only to match a gold contract exactly, but to infer a contract that is useful for downstream tool filtering. A learned contract may be imperfect while still preserving enough causal structure to improve agent reliability.

\subsection{Intrinsic Accuracy and Downstream Utility}
We distinguish two evaluation settings. Intrinsic evaluation measures whether the predicted contract matches the gold contract. For preconditions and effects, this can be measured using set precision, recall, and F1 against the gold fields. For risk labels, this can be measured using classification accuracy.

Downstream evaluation measures whether learned contracts support reliable execution when used by a causal tool-filtering method. Let $\mathcal{C} = \{c_i\}_{i=1}^{n}$ denote the gold contract set and $\hat{\mathcal{C}} = \{\hat{c}_i\}_{i=1}^{n}$ denote the learned contract set. The downstream question is whether an agent using $\hat{\mathcal{C}}$ preserves the reliability and efficiency of an agent using $\mathcal{C}$.

Contract errors can affect filtering in two main ways. Over-filtering occurs when learned contracts are too restrictive and hide a tool needed for the next step. Under-filtering occurs when learned contracts are too permissive and expose tools that are irrelevant, premature, or non-goal-directed. Contract2Tool is evaluated by measuring both intrinsic contract quality and downstream agent behavior under these learned contracts.

% =========================================================
% 4. Contract2Tool Method
% =========================================================
\section{Contract2Tool Method}
\label{sec:method}

\begin{figure}[t]
\centering
\begin{tikzpicture}[node distance=6mm]
\node[procbox] (evidence)
{Tool evidence\\
(name, description, schema,\\
documentation, traces)};

\node[procbox, below=of evidence] (generator)
{Contract generator};

\node[procbox, below=of generator] (raw)
{Raw contract prediction\\
(requires, produces, risk, cost)};

\node[procbox, below=of raw] (norm)
{Normalization and validation\\
(canonicalization, validation,\\
duplicate removal)};

\node[accentbox, below=of norm] (learned)
{Learned tool contract\\
(canonical preconditions,\\
effects, risk, cost)};

\draw[flow] (evidence) -- (generator);
\draw[flow] (generator) -- (raw);
\draw[flow] (raw) -- (norm);
\draw[flow] (norm) -- (learned);
\end{tikzpicture}
\caption{Overview of Contract2Tool. Observable tool evidence is converted into a normalized symbolic contract through generation, normalization, and validation before being used by downstream causal tool-filtering methods.}
\label{fig:contract2tool_pipeline}
\end{figure}
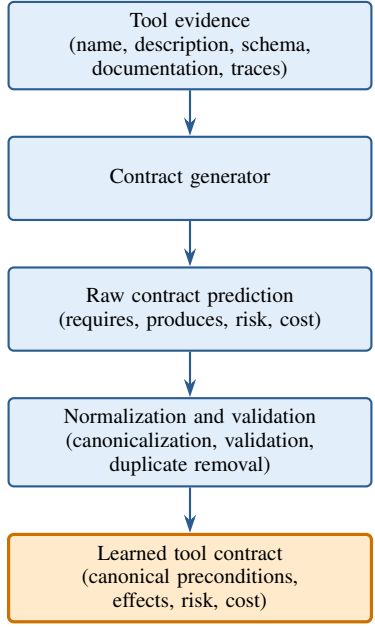

Contract2Tool converts observable tool evidence into normalized symbolic contracts for downstream tool filtering. As summarized in Figure~\ref{fig:contract2tool_pipeline}, the pipeline has four stages: constructing an evidence view for each tool, generating a raw contract prediction, normalizing and validating the predicted fields, and producing a learned contract set for downstream filtering. Execution traces can be included as part of the evidence view when available. The final output is a learned contract set
\begin{equation}
\hat{\mathcal{C}} = \{\hat{c}_i \mid i = 1,\ldots,n\}.
\end{equation}
This learned contract set can be used in place of manually specified contracts.

\subsection{Evidence Conditions}

Contract2Tool operates under different levels of available tool evidence. We evaluate progressively richer evidence conditions to measure how additional information affects contract quality:
\begin{itemize}
\item \textit{Name}: the tool name alone.
\item \textit{Metadata}: the tool name and natural-language description.
\item \textit{Schema}: metadata plus input/output schema.
\item \textit{Documentation}: metadata and schema plus short API-style documentation.
\item \textit{Traces}: execution traces containing before-and-after task states, tool outputs, and execution status.
\item \textit{Hybrid}: documentation and traces combined.
\end{itemize}
These conditions reflect realistic levels of information available in tool ecosystems. Some systems expose only names and schemas, while mature systems may also maintain documentation and execution logs. The framework can incorporate both successful and failed traces when available; in the experiments, we primarily use controlled successful task trajectories to expose produced state variables through before-and-after state changes.

\subsection{Contract Generation}

Given an evidence view $z_i^{(m)}$ for tool $t_i$ under condition $m$, a contract generator $f_\theta$ produces a structured contract prediction:
\begin{equation}
f_\theta(z_i^{(m)}) = \hat{c}_i.
\end{equation}
The predicted contract contains preconditions, effects, risk, and optional cost fields. For example, a calendar update tool may require an event identifier, produce an updated-event state, and receive a high risk label because it modifies external state. The \texttt{requires} and \texttt{produces} fields are interpreted as task-state variables. The \texttt{risk} field is selected from a fixed label set, such as low, medium, or high. The \texttt{cost} field is optional and may represent coarse latency, token, or execution cost. Rationales may be generated for inspection, but downstream filtering uses only the structured contract fields.

\subsection{Normalization and Validation}

Generated contracts may contain synonyms, invalid variables, duplicate entries, or malformed outputs. Contract2Tool therefore applies a normalization and validation layer before using a predicted contract. This layer assumes a predefined canonical state-variable vocabulary for the benchmark or deployment environment. Normalization maps generated fields into this vocabulary. For example, surface forms such as ``event identifier'' or ``document text'' are mapped to canonical state variables in the benchmark vocabulary. Validation checks that required and produced variables belong to the allowed vocabulary, risk labels are valid, duplicate variables are removed, and malformed outputs are rejected or marked invalid. This step is important because downstream causal filtering depends on exact symbolic state variables.

\subsection{Trace Evidence}

Execution traces provide direct evidence about tool-induced state changes. A trace records the state before a tool call, the selected tool, the arguments, execution status, output observation, and state after execution:
\begin{equation}
(s_{\mathrm{before}}, t_i, \mathrm{args}, \mathrm{status}, o_i, s_{\mathrm{after}}).
\end{equation}
Successful traces reveal produced state variables through before-and-after state differences. Failed traces, when available, can suggest missing preconditions. In the experiments in this paper, we focus on controlled successful trajectories and use state differences to identify produced effects. This is useful because documentation may describe a human-facing output, such as a list of search results, while downstream filtering may require a more specific state variable, such as an identifier needed by the next tool.

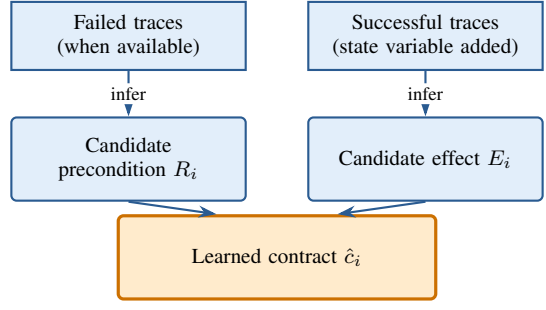
\begin{figure}[t]
\centering
\begin{tikzpicture}[node distance=6mm and 8mm]

% Top trace evidence nodes
\node[databox, minimum width=31mm, text width=27mm] (fail)
{Failed traces\
(when available)};

\node[databox, minimum width=31mm, text width=27mm, right=of fail] (success)
{Successful traces\
(state variable added)};

% Candidate inference nodes
\node[procbox, minimum width=31mm, text width=27mm, below=of fail] (precond)
{Candidate\
precondition $R_i$};

\node[procbox, minimum width=31mm, text width=27mm, below=of success] (effect)
{Candidate\
effect $E_i$};

% Refined learned contract node
\node[accentbox, minimum width=42mm, text width=38mm]
(refined) at ($(precond)!0.5!(effect)+(0,-13mm)$)
{Learned contract\
$\hat{c}_i$};

% Arrows
\draw[flow] (fail) -- (precond) node[edgelbl] {infer};
\draw[flow] (success) -- (effect) node[edgelbl] {infer};

\draw[flow] (precond.south) -- ([xshift=-8mm]refined.north);
\draw[flow] (effect.south) -- ([xshift=8mm]refined.north);

\end{tikzpicture}
\caption{Use of trace evidence in contract learning. Successful traces reveal produced state variables through before-and-after state differences, while failed traces, when available, can suggest missing preconditions.}
\label{fig:trace_refinement}
\end{figure}

\subsection{Learned Contract Set}

After generation, normalization, validation, and optional use of trace evidence, Contract2Tool returns a learned contract set:
\begin{equation}
\hat{\mathcal{C}} = \{\hat{c}_1, \hat{c}_2, \ldots, \hat{c}_n\}.
\end{equation}
This learned contract set can be evaluated intrinsically against gold contracts or used downstream by causal tool-filtering methods. The downstream question is whether $\hat{\mathcal{C}}$ preserves the causal edges needed to expose the correct next tool during multi-step execution, even when individual contract fields are imperfect.

% =========================================================
% 5. Benchmark Design
% =========================================================
\section{Benchmark Design}
\label{sec:benchmark}

We design the benchmark to evaluate Contract2Tool at three levels. First, we measure whether learned contracts match gold preconditions, effects, and risk labels. Second, we use a filter-oracle evaluation to test whether learned contracts expose the gold next tool under causal filtering. Third, we measure whether learned contracts preserve downstream reliability when used inside multi-step agents. This separates intrinsic contract quality, filter-level causal utility, and downstream agent behavior.

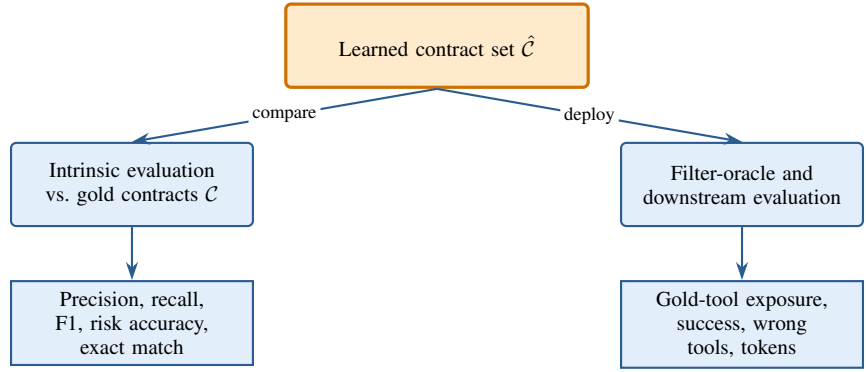
\begin{figure}[t]
\centering
\begin{tikzpicture}[node distance=7mm and 8mm]

% Learned contract set
\node[accentbox, minimum width=40mm, text width=36mm] (learned)
{Learned contract set\
$\hat{\mathcal{C}}$};

% Evaluation branches
\node[procbox, minimum width=32mm, text width=28mm,
below left=of learned, xshift=4mm] (intrinsic)
{Intrinsic evaluation\
vs.\ gold contracts\
$\mathcal{C}$};

\node[procbox, minimum width=32mm, text width=28mm,
below right=of learned, xshift=-4mm] (downstream)
{Filter-oracle\
and downstream\
evaluation};

% Metric nodes
\node[databox, minimum width=32mm, text width=28mm,
below=of intrinsic] (imetrics)
{Precision, recall, F1,\
risk accuracy,\
exact match};

\node[databox, minimum width=32mm, text width=28mm,
below=of downstream] (dmetrics)
{Gold-tool exposure,\
success, wrong tools,\
tokens};

% Arrows
\draw[flow] (learned.south) -- (intrinsic.north) node[edgelbl] {compare};
\draw[flow] (learned.south) -- (downstream.north) node[edgelbl] {deploy};
\draw[flow] (intrinsic) -- (imetrics);
\draw[flow] (downstream) -- (dmetrics);

\end{tikzpicture}
\caption{Evaluation tracks for Contract2Tool. Learned contracts are evaluated intrinsically against gold contracts and downstream through filter-oracle and agent-task evaluations.}
\label{fig:evaluation_tracks}
\end{figure}

\subsection{Tool Registry and Gold Contracts}

We build on a controlled synthetic tool-use benchmark with a registry of 100 tools. The registry spans workflow domains such as calendar, email, files/documents, and controlled distractor domains. Each tool includes a name, natural-language description, input/output schema, optional API-style documentation, gold preconditions, gold effects, a risk label, and an optional cost label. The registry includes task-relevant tools, near-duplicate tools, semantically plausible distractors, premature tools, risky tools, and cross-domain distractors. These distractors make contract learning consequential: missing or spurious preconditions and effects can cause downstream over-filtering or under-filtering.

\subsection{Contract-Learning Examples}

For each tool $t_i$, we construct contract-learning examples by pairing an evidence view with the gold contract target:
\begin{equation}
z_i^{(m)} \rightarrow c_i = (R_i, E_i, \rho_i, k_i),
\end{equation}
where $m$ denotes the evidence condition. Each evidence view hides the gold contract fields and exposes only the information available under that condition. We evaluate name-only, metadata, schema, documentation, trace, and hybrid documentation-plus-trace views.

\begin{table}[t]
\centering
\footnotesize
\caption{Benchmark components for evaluating Contract2Tool intrinsically and downstream.}
\label{tab:benchmark_components}
\begin{tabularx}{\linewidth}{@{}lX@{}}
\toprule
Component & Description \tabularnewline
\midrule
Tool registry & 100 synthetic tools with names, descriptions, schemas, documentation, gold contracts, risk labels, and optional cost labels. \tabularnewline
Evidence views & Name, metadata, schema, documentation, traces, and hybrid documentation-plus-trace views. \tabularnewline
Contract targets & Gold preconditions, effects, risk labels, and optional cost labels. \tabularnewline
Execution traces & Controlled mocked tool executions with before-and-after task states; experiments primarily use successful task trajectories. \tabularnewline
Downstream tasks & 102 multi-step agent tasks across calendar, email, and file/document workflows. \tabularnewline
\bottomrule
\end{tabularx}
\end{table}

\subsection{Execution Traces}

We generate controlled execution traces from the benchmark environment. Each trace records the state before execution, selected tool, arguments, execution status, mocked output, and state after execution. In the experiments reported here, trace evidence primarily comes from successful gold task trajectories, which reveal produced state variables through before-and-after state differences. The framework can also incorporate failed traces when available, which may provide evidence about missing preconditions.

\subsection{Filter-Oracle Evaluation}

Intrinsic contract quality does not fully determine whether a learned contract is useful for causal filtering. We therefore include a filter-oracle evaluation that uses the gold task trajectory to isolate contract and filter behavior from model-specific tool-calling behavior. At each gold decision step, CMTF is run using a candidate contract set, and we measure whether the filter exposes the gold next tool. We report gold-tool exposure, no-visible rate, average visible tools, and extra tools exposed.

\subsection{Downstream Agent Tasks}

For downstream evaluation, we reuse the multi-step tool-use benchmark used for causal tool filtering. The benchmark contains 102 tasks across calendar, email, and file/document workflows. The same task set is executed under five filtering methods: all-tools exposure, keyword top-5 filtering, state-aware filtering, gold-contract CMTF, and learned-contract CMTF. This setup allows us to measure how much reliability learned contracts preserve relative to gold-contract causal filtering and how much tool exposure they reduce relative to non-causal baselines.

% =========================================================
% 6. Experimental Setup
% =========================================================
\section{Experimental Setup}
\label{sec:experimental_setup}

We evaluate Contract2Tool with three complementary evaluations. First, we measure intrinsic contract quality by comparing learned contracts against gold preconditions, effects, and risk labels. Second, we use a filter-oracle evaluation to test whether learned contracts expose the gold next tool under causal filtering. Third, we measure downstream utility by using learned contracts inside causal tool filtering for multi-step agent tasks. This design tests whether learned contracts are merely plausible, whether they preserve the causal frontier needed by the filter, and whether they improve tool-use behavior in downstream execution.

\begin{table}[t]
\centering
\footnotesize
\caption{Summary of the experimental setup for intrinsic, filter-oracle, and downstream evaluations.}
\label{tab:experimental_setup}
\begin{tabularx}{\linewidth}{@{}lX@{}}
\toprule
Component & Setting \tabularnewline
\midrule
Contract generators & Instruction-following LLMs evaluated under multiple evidence conditions. \tabularnewline
Evidence conditions & Name, metadata, schema, documentation, traces, and hybrid evidence. \tabularnewline
Intrinsic metrics & Preconditions/effects precision, recall, and F1; risk accuracy; exact match; invalid-output rate. \tabularnewline
Filter-oracle metrics & Gold-tool exposure, no-visible rate, average visible tools, and extra tools exposed. \tabularnewline
Downstream comparisons & All-tools exposure, keyword top-5, state-aware filtering, gold-contract CMTF, and learned-contract CMTF. \tabularnewline
Downstream metrics & Task success, wrong-tool count, premature-action count, average tools exposed per step, and token cost. \tabularnewline
\bottomrule
\end{tabularx}
\end{table}

\subsection{Contract Inference Protocol}

Each contract generator receives one evidence view for a tool and outputs a structured contract with \texttt{requires}, \texttt{produces}, \texttt{risk}, and optional \texttt{cost} fields. LLM-based generators infer contracts under each evidence condition using a fixed output schema. All generated contracts are normalized and validated before scoring. Invalid outputs, unresolved variables, and normalization failures are recorded separately so that contract quality is not conflated with formatting quality.

The intrinsic experiments evaluate multiple Bedrock-hosted model families, including Amazon Nova, Anthropic Claude, and Meta Llama models. For the main downstream learned-contract run, we use the strongest learned contract set: Claude Opus 4.8 with hybrid documentation-and-trace evidence.

\subsection{Evidence Conditions}

We evaluate progressively richer evidence conditions:
\begin{itemize}
\item \textbf{Name:} the tool name alone.
\item \textbf{Metadata:} tool name and natural-language description.
\item \textbf{Schema:} metadata plus input/output schema.
\item \textbf{Documentation:} metadata and schema plus short API-style documentation.
\item \textbf{Traces:} controlled execution traces containing before-and-after task states, tool outputs, and execution status.
\item \textbf{Hybrid:} documentation and traces combined.
\end{itemize}
The framework can incorporate failed traces when available; the experiments reported here primarily use successful controlled task trajectories to expose produced state variables through before-and-after state changes.

\subsection{Intrinsic Contract Metrics}

For preconditions and effects, we report set precision, recall, and F1 against the gold contract fields. For risk labels, we report classification accuracy. We also report exact contract match, invalid-output rate, and normalization failure rate. These metrics measure whether the generated contracts recover the symbolic structure needed for causal filtering.

\subsection{Filter-Oracle Metrics}

Intrinsic metrics do not fully determine whether a learned contract set is useful for causal filtering. We therefore include a filter-oracle evaluation that isolates contract and filter behavior from model-specific tool-calling behavior. At each gold decision step, CMTF is run using a candidate contract set, and we measure whether the filter exposes the gold next tool. We report gold-tool exposure, no-visible rate, average visible tools, and extra tools exposed.

\subsection{Downstream Agent Metrics}

For downstream evaluation, we run the same 102 multi-step tasks under five filtering methods: all-tools exposure, keyword top-5 filtering, state-aware filtering, gold-contract CMTF, and learned-contract CMTF. We report task success, wrong-tool count, premature-action count, average tools exposed per step, and token cost. The main downstream aggregate is computed over the four reliable tool-calling models. Meta Llama 3.3 70B is reported separately because it exhibited tool-call compatibility failures across all methods, including gold-contract CMTF.

% =========================================================
% 7. Results
% =========================================================
\section{Results}
\label{sec:results}

We evaluate Contract2Tool along three dimensions: intrinsic contract quality, filter-level causal utility, and downstream agent performance. Intrinsic evaluation tests whether learned contracts recover gold preconditions, effects, and risk labels. The filter-oracle evaluation isolates whether learned contracts expose the gold next tool under causal filtering. Downstream evaluation tests whether learned contracts preserve task success when deployed inside CMTF.

\subsection{Intrinsic Contract Quality}

Table~\ref{tab:intrinsic_by_evidence} reports average contract-learning performance across the initial evidence conditions. Schema and documentation provide the strongest precondition recovery, reaching 0.99 and 1.00 Requires F1, respectively. Name-only and metadata-only settings recover useful effect information, but are less reliable for exact contract recovery. This suggests that tool names and descriptions often reveal what a tool is about, while schemas and documentation are more useful for identifying the state variables required for reliable execution.

\begin{table}[!t]
\centering
\caption{Average intrinsic contract-learning performance across evidence conditions.}
\label{tab:intrinsic_by_evidence}
\scriptsize
\begin{tabular}{@{}lcccc@{}}
\toprule
Evidence & Req. F1 & Prod. F1 & Risk & Exact \tabularnewline
\midrule
Name & 0.77 & 0.86 & 0.63 & 0.43 \tabularnewline
Metadata & 0.72 & 0.86 & 0.67 & 0.44 \tabularnewline
Schema & 0.99 & 0.80 & 0.67 & 0.54 \tabularnewline
Documentation & 1.00 & 0.77 & 0.67 & 0.51 \tabularnewline
\bottomrule
\end{tabular}
\end{table}

\subsection{Trace and Hybrid Evidence}

Documentation and schemas improve intrinsic contract quality, but they can still miss task-critical causal effects. For example, several models inferred that a calendar search tool produces a list of events, while the benchmark requires a specific event identifier in order to continue to event update or event read operations. Execution traces expose these before-and-after state transitions directly.

Trace and hybrid evidence improve contract recovery for the strongest generators. Hybrid evidence gives the best overall results: Claude Opus 4.8 reaches 1.00 Requires F1, 0.98 Produces F1, and 0.77 exact match, while Claude Sonnet 4.6 reaches 1.00 Requires F1, 0.99 Produces F1, and 0.74 exact match. These results suggest that traces are especially useful when static documentation describes a human-facing output but the downstream workflow depends on a more specific state variable.

\subsection{Filter-Oracle Evaluation}

Intrinsic accuracy does not fully determine whether a learned contract is useful for causal filtering. A contract can have high average F1 while still missing a causal edge required by a task. We therefore evaluate whether each learned contract set exposes the gold next tool at each step when used inside CMTF, assuming the task follows the gold trajectory.

Table~\ref{tab:filter_oracle} shows that documentation- and schema-derived Opus 4.8 contracts expose the gold next tool for 84.1\% of decision steps and produce no visible tool in 7.8\% of steps. In contrast, trace and hybrid contracts recover the gold causal frontier exactly in this oracle evaluation. Opus 4.8 hybrid and Sonnet 4.6 hybrid both achieve 1.00 gold exposure, 0.00 no-visible rate, and one visible tool per step.

\begin{table}[!t]
\centering
\caption{Filter-oracle evaluation over 283 gold decision steps. Gold exp. denotes gold-tool exposure; No vis. denotes no visible tool.}
\label{tab:filter_oracle}
\scriptsize
\begin{tabular}{@{}lcccc@{}}
\toprule
Source & Gold exp. & No vis. & Tools & Extra \tabularnewline
\midrule
Gold & 1.000 & 0.000 & 1.000 & 0.000 \tabularnewline
Opus 4.8 Docs & 0.841 & 0.078 & 0.922 & 0.081 \tabularnewline
Opus 4.8 Schema & 0.841 & 0.078 & 0.922 & 0.081 \tabularnewline
Opus 4.8 Traces & 1.000 & 0.000 & 1.000 & 0.000 \tabularnewline
Opus 4.8 Hybrid & 1.000 & 0.000 & 1.000 & 0.000 \tabularnewline
Sonnet 4.6 Hybrid & 1.000 & 0.000 & 1.000 & 0.000 \tabularnewline
Llama 3.3 Docs & 0.604 & 0.000 & 1.000 & 0.396 \tabularnewline
Llama 3.3 Schema & 0.604 & 0.000 & 1.000 & 0.396 \tabularnewline
\bottomrule
\end{tabular}
\end{table}

\subsection{Downstream Agent Performance}

Finally, we evaluate whether learned contracts preserve downstream agent performance. We use the strongest learned contract source, Claude Opus 4.8 with hybrid documentation-and-trace evidence, inside CMTF. Table~\ref{tab:downstream_aggregate_non_llama} reports aggregate performance over the four reliable tool-calling models: Nova 2 Lite, Nova 2 Pro, Claude 3.5 Haiku, and Claude Sonnet 4.6. We exclude Llama 3.3 70B from the main aggregate because it showed near-zero success even under gold-contract CMTF, suggesting a tool-call compatibility issue rather than a learned-contract failure.

Learned CMTF nearly matches gold-contract CMTF. Gold CMTF achieves 0.990 success, while learned CMTF achieves 0.980 success. Learned CMTF also reduces wrong-tool calls to 0.020 per task, eliminates premature actions, and exposes exactly one tool per step. Relative to all-tools exposure, learned CMTF reduces visible tools from 100.0 to 1.0 and reduces average token usage from 26,172 to 2,528 tokens per task.

\begin{table}[!t]
\centering
\caption{Aggregate downstream performance over the four reliable tool-calling models using Opus 4.8 hybrid learned contracts.}
\label{tab:downstream_aggregate_non_llama}
\scriptsize
\begin{tabular}{@{}lccccc@{}}
\toprule
Method & Success & Wrong & Prem. & Tools & Tokens \tabularnewline
\midrule
All tools & 0.775 & 1.461 & 0.027 & 100.000 & 26172 \tabularnewline
Keyword top-5 & 0.620 & 2.265 & 0.032 & 5.000 & 4482 \tabularnewline
State-aware & 0.620 & 2.196 & 0.000 & 5.733 & 4648 \tabularnewline
Gold CMTF & 0.990 & 0.010 & 0.000 & 1.000 & 2532 \tabularnewline
Learned CMTF & 0.980 & 0.020 & 0.000 & 1.000 & 2528 \tabularnewline
\bottomrule
\end{tabular}
\end{table}

At the model level, learned CMTF exactly matches gold CMTF for Nova 2 Lite, Nova 2 Pro, and Claude Sonnet 4.6. For Claude 3.5 Haiku, learned CMTF is slightly lower than gold CMTF, with success decreasing from 0.96 to 0.92 and wrong-tool calls increasing from 0.04 to 0.08 per task. Even in this case, learned CMTF substantially outperforms all-tools, keyword top-5, and state-aware filtering.

% =========================================================
% 8. Discussion
% =========================================================
\section{Discussion}
\label{sec:discussion}

\subsection{Contracts as the Missing Layer Between Schemas and Agents}

The results support the view that tool contracts form a useful intermediate layer between tool schemas and agent execution. Schemas describe how to call a tool, including its arguments and input format. Contracts describe when the tool should be exposed and what task-state transition it enables. This distinction matters for multi-step tool use because an agent may know how to call a tool while still calling it too early, selecting a semantically related but unnecessary tool, or failing to expose the next tool required by the workflow.

\subsection{Why Trace Evidence Matters}

Documentation and schemas are useful, but they do not always describe the state variables needed for causal filtering. A tool description may say that a calendar search returns a list of events, while the downstream workflow may require a specific event identifier before an update or read operation can proceed. This distinction appeared in the filter-oracle evaluation: documentation- and schema-derived contracts exposed the gold next tool in 84.1\% of steps, while trace and hybrid contracts recovered the gold causal frontier exactly. Execution traces expose these before-and-after state changes directly, allowing learned contracts to observe which state variables appear after successful tool execution instead of relying only on static tool descriptions.

\subsection{Intrinsic Accuracy Is Necessary but Not Sufficient}

The experiments show that intrinsic contract metrics alone do not fully determine downstream usefulness. A contract set can achieve high average precondition or effect F1 while still missing a task-critical causal edge. Such errors can cause over-filtering, where the needed next tool is hidden, or under-filtering, where extra irrelevant tools remain visible. The filter-oracle evaluation helps address this gap by testing whether learned contracts preserve the gold causal frontier step by step. This intermediate evaluation separates contract quality from model-specific tool-calling behavior. It also shows that exact contract match is a conservative metric: it can penalize harmless extra fields and abstraction differences, while downstream filtering depends most on preserving task-critical causal edges.

\subsection{Learned Contracts Can Approximate Gold-Contract Filtering}

The downstream results show that learned contracts can recover most of the practical value of manually written contracts. Using contracts inferred from hybrid documentation-and-trace evidence, learned CMTF nearly matches gold-contract CMTF in task success while preserving the same minimal tool exposure. In the downstream aggregate, learned CMTF achieves 0.980 success compared with 0.990 for gold-contract CMTF, while maintaining the same average exposure of one visible tool per step. This suggests that exact contract recovery is not always required. What matters most is preserving the causal structure needed to expose the next useful tool and hide premature or irrelevant tools.

\subsection{Toward Self-Maintaining Tool Registries}

Large tool ecosystems change over time as APIs are added, deprecated, renamed, or modified. Manually maintaining preconditions, effects, and risk annotations for every tool is difficult and error-prone. Contract2Tool suggests a path toward self-maintaining tool registries, where metadata, schemas, documentation, and execution traces are used to bootstrap and refresh tool contracts. In practice, such systems should remain human-reviewable, especially for high-risk actions such as sending, deleting, sharing, or externally modifying state. Risk annotations are especially important to review because risk is context-dependent and may depend on user intent, permissions, and organizational policy. Learned contracts are best viewed as a scalable contract-generation and monitoring layer rather than an unchecked replacement for safety review.

% =========================================================
% 9. Limitations and Threats to Validity
% =========================================================
\section{Limitations and Threats to Validity}
\label{sec:limitations}

This study uses a controlled synthetic tool registry with gold contracts. This design allows us to isolate contract learning and causal filtering behavior, but it may not capture the full complexity of real API ecosystems. Production tools often have incomplete documentation, inconsistent schemas, changing behavior, authentication constraints, rate limits, and side effects that are difficult to model in a controlled benchmark.

The experiments also assume a fixed canonical vocabulary of task-state variables. In practice, defining this vocabulary is itself a design challenge. Contract labels may depend on the chosen abstraction level. For example, a search tool may reasonably be described as producing a list of results, a specific identifier, or both. As a result, gold contracts should be understood as benchmark-specific operational labels rather than universal semantic truths.

Trace evidence is valuable because it reveals before-and-after state transitions, but our trace experiments primarily use controlled mocked task trajectories. Real execution traces may be sparse, noisy, biased toward common workflows, or missing rare but important tool behaviors. A tool may have valid preconditions or effects that never appear in the available traces. Future work should evaluate Contract2Tool on larger collections of production-style traces, including failed executions and changing API behavior.

The downstream evaluation uses mocked tool execution to isolate tool-selection behavior. This lets us measure whether learned contracts preserve the causal filtering benefits of gold contracts, but it does not fully test real API failures, latency variation, authorization errors, nondeterministic outputs, or irreversible side effects. High-risk tools such as sending, deleting, sharing, or externally modifying state require additional safeguards before learned contracts are used in production systems.

Finally, downstream results can be affected by model-specific tool-calling behavior. Some models may fail because of formatting, tool-call compatibility, or API invocation issues rather than because of contract quality. We therefore interpret downstream comparisons as measuring the combined behavior of the learned contract set, the filtering method, and the tool-calling model. Broader validation on real tools, larger registries, and additional model families remains an important direction for future work.

% =========================================================
% 10. Conclusion
% =========================================================
\section{Conclusion}
\label{sec:conclusion}

This paper introduced Contract2Tool, a framework for learning lightweight tool contracts for tool-augmented LLM agents. The central motivation is that causal tool filtering depends on preconditions, effects, and risk annotations, but manually defining and maintaining these contracts does not scale to large or changing tool ecosystems. Contract2Tool addresses this gap by inferring structured contracts from tool metadata, schemas, documentation, and execution traces, then evaluating them both intrinsically and downstream inside causal filtering.

The results show that learned contracts can preserve most of the reliability and efficiency benefits of gold contracts. Using contracts inferred from hybrid documentation-and-trace evidence, learned CMTF achieved 0.980 downstream success, close to 0.990 for gold-contract CMTF, while reducing visible tools from 100 to 1 and reducing average token usage from 26,172 to 2,528 relative to all-tools exposure. The results also show that trace evidence is important because it exposes task-state transitions that schemas and documentation may omit.

More broadly, this work suggests that reliable tool-augmented agents need a contract layer between tool schemas and agent execution. Schemas describe how to call tools, while contracts describe when tools should be exposed and what state changes they produce. Future work should validate contract learning on real API ecosystems, larger dynamic tool registries, production execution traces, and higher-risk tool-use settings.

% =========================================================
% Acknowledgments
% =========================================================
\section*{Acknowledgment}
The authors thank colleagues for helpful feedback. This work was conducted in the authors' personal capacity. The views expressed in this paper are solely those of the authors and do not necessarily reflect the views of their employers. This work did not receive external funding. The authors declare no conflicts of interest.

% =========================================================
% Artifact Availability
% =========================================================
\section*{Artifact Availability}
The synthetic benchmark, tool registry, filtering implementations, evaluation scripts, and analysis utilities are available at: \url{https://github.com/R-Suresh/Contract2Tool}.

% =========================================================
% References
% =========================================================
\bibliographystyle{IEEEtran}
\bibliography{references}

@misc{babu2026selfhealing,
  title         = {Self-Healing Agentic Orchestrators for Reliable Tool-Augmented Large Language Model Systems},
  author        = {Babu, Rahul Suresh and Agrawal, Adarsh},
  year          = {2026},
  eprint        = {2606.01416},
  archivePrefix = {arXiv},
  primaryClass  = {cs.AI},
  url           = {https://arxiv.org/abs/2606.01416}
}

@misc{babu2026toolchoice,
  title         = {{ToolChoiceConfusion}: Causal Minimal Tool Filtering for Reliable {LLM} Agents},
  author        = {Babu, Rahul Suresh and Iyer, Laxmipriya Ganesh},
  year          = {2026},
  eprint        = {2606.06284},
  archivePrefix = {arXiv},
  primaryClass  = {cs.AI},
  url           = {https://arxiv.org/abs/2606.06284}
}

@article{fikes1971strips,
  title={STRIPS: A New Approach to the Application of Theorem Proving to Problem Solving},
  author={Fikes, Richard E. and Nilsson, Nils J.},
  journal={Artificial Intelligence},
  volume={2},
  number={3--4},
  pages={189--208},
  year={1971}
}

@inproceedings{li2023apibank,
  title={API-Bank: A Comprehensive Benchmark for Tool-Augmented LLMs},
  author={Li, Minghao and Zhao, Yingxiu and Yu, Bowen and Song, Feifan and Li, Hangyu and Yu, Haiyang and Li, Zhoujun and Huang, Fei and Li, Yongbin},
  booktitle={Proceedings of the 2023 Conference on Empirical Methods in Natural Language Processing},
  year={2023},
  url={https://arxiv.org/abs/2304.08244}
}

@misc{liu2025toolscope,
  title={ToolScope: Enhancing LLM Agent Tool Use through Tool Merging and Context-Aware Filtering},
  author={Liu, Marianne Menglin and Garcia, Daniel and Parllaku, Fjona and Upadhyay, Vikas and Shah, Syed Fahad Allam and Roth, Dan},
  year={2025},
  note={arXiv preprint},
  eprint={2510.20036},
  archivePrefix={arXiv},
  primaryClass={cs.CL},
  url={https://arxiv.org/abs/2510.20036}
}

@techreport{mcdermott1998pddl,
  title={{PDDL}: The Planning Domain Definition Language},
  author={McDermott, Drew and Ghallab, Malik and Howe, Adele and Knoblock, Craig and Ram, Ashwin and Veloso, Manuela and Weld, Daniel and Wilkins, David},
  institution={Yale Center for Computational Vision and Control},
  year={1998}
}

@inproceedings{patil2024bfcl,
  title={The Berkeley Function-Calling Leaderboard},
  author={Patil, Shishir G. and Zhang, Tianjun and Wang, Xin and Gonzalez, Joseph E.},
  booktitle={Proceedings of Machine Learning Research},
  year={2025},
  url={https://proceedings.mlr.press/v267/patil25a.html}
}

@inproceedings{qin2023toollm,
  title={ToolLLM: Facilitating Large Language Models to Master 16000+ Real-world APIs},
  author={Qin, Yujia and Liang, Shihao and Ye, Yining and Zhu, Kunlun and Yan, Lan and Lu, Yaxi and Lin, Yankai and Cong, Xin and Tang, Xiangru and Qian, Bill and Zhao, Sihan and Tian, Runchu and Xie, Ruobing and Zhou, Jie and Gerstein, Mark and Li, Dahai and Liu, Zhiyuan and Sun, Maosong},
  booktitle={International Conference on Learning Representations},
  year={2024},
  url={https://arxiv.org/abs/2307.16789}
}

@inproceedings{schick2023toolformer,
  title={Toolformer: Language Models Can Teach Themselves to Use Tools},
  author={Schick, Timo and Dwivedi-Yu, Jane and Dess{\`i}, Roberto and Raileanu, Roberta and Lomeli, Maria and Hambro, Eric and Zettlemoyer, Luke and Cancedda, Nicola and Scialom, Thomas},
  booktitle={Advances in Neural Information Processing Systems},
  year={2023},
  url={https://arxiv.org/abs/2302.04761}
}

@inproceedings{shi2025toolret,
  title={Retrieval Models Aren't Tool-Savvy: Benchmarking Tool Retrieval for Large Language Models},
  author={Shi, Zhengliang and Wang, Yuhan and Yan, Lingyong and Ren, Pengjie and Wang, Shuaiqiang and Yin, Dawei and Ren, Zhaochun},
  booktitle={Findings of the Association for Computational Linguistics},
  year={2025},
  url={https://arxiv.org/abs/2503.01763}
}

@inproceedings{yao2022react,
  title={ReAct: Synergizing Reasoning and Acting in Language Models},
  author={Yao, Shunyu and Zhao, Jeffrey and Yu, Dian and Du, Nan and Shafran, Izhak and Narasimhan, Karthik and Cao, Yuan},
  booktitle={International Conference on Learning Representations},
  year={2023},
  url={https://arxiv.org/abs/2210.03629}
}

\end{document}